\documentclass{article} 
\usepackage{nips15submit_e,times}
\usepackage{hyperref}
\usepackage{url}

\usepackage{algorithm}
\usepackage{algorithmic}

\usepackage{epsf}
\usepackage{epsfig}
\usepackage{multirow}
\usepackage{tikz}
\usepackage{mathtools}
\usepackage{bm}
\usepackage{amsmath, amsthm, amssymb}
\usepackage{bbm}
\usepackage{wrapfig}
\usepackage{enumitem}

\usepackage[utf8]{inputenc} 
\usepackage[T1]{fontenc}    
\usepackage{booktabs}       
\usepackage{amsfonts}       
\usepackage{nicefrac}       
\usepackage{microtype}      
\usepackage{color}

\usepackage{graphicx} 
\usepackage{float}
\usepackage{stackengine}
\usepackage{caption}
\usepackage{subcaption}
\usepackage{multirow}
\usepackage{array}
\usepackage{wrapfig} 

\usepackage{algorithm}
\usepackage{algorithmic}

\usepackage{amsmath}
\usepackage{amssymb}
\usepackage{mathrsfs}
\usepackage{bm}
\usepackage{tabu}

\newcolumntype{L}[1]{>{\raggedright\let\newline\\\arraybackslash\hspace{0pt}}m{#1}}
\newcolumntype{C}[1]{>{\centering\let\newline\\\arraybackslash\hspace{0pt}}m{#1}}
\newcolumntype{R}[1]{>{\raggedleft\let\newline\\\arraybackslash\hspace{0pt}}m{#1}}

\newcommand{\comment}[1]{}

\long\def\ignore#1{}

\nipsfinalcopy

\title{VisualBackProp for learning using privileged information with CNNs}

\author{
  Devansh Bisla \\
  Department of Electrical Engineering, \\
 NYU Tandon School of Engineering\\
  \texttt{db3484@nyu.edu} \\
  \And
  Anna Choromanska \\
  Department of Electrical Engineering, \\
 NYU Tandon School of Engineering, \\
  \texttt{ac5455@nyu.edu} \\
}

\begin{document}

\maketitle

\begin{abstract}

In many machine learning applications, from medical diagnostics to autonomous driving, the availability of prior knowledge can be used to improve the predictive performance of learning algorithms and incorporate ``physical,'' ``domain knowledge,'' or ``common sense'' concepts into training of machine learning systems as well as verify constraints/properties of the systems. We explore the learning using privileged information paradigm and show how to incorporate the privileged information, such as segmentation mask available along with the classification label of each example, into the training stage of convolutional neural networks. This is done by augmenting the CNN model with an architectural component that effectively focuses model's attention on the desired region of the input image during the training process and that is transparent to the network's label prediction mechanism at testing. This component effectively corresponds to the visualization strategy for identifying the parts of the input, often referred to as visualization mask, that most contribute to the prediction, yet uses this strategy in reverse to the classical setting in order to enforce the desired visualization mask instead. We verify our proposed algorithms through exhaustive experiments on benchmark ImageNet and PASCAL VOC data sets and achieve improvements in the performance of $2.4\%$ and $2.7\%$ over standard single-supervision model training. Finally, we confirm the effectiveness of our approach on skin lesion classification problem.
\end{abstract}

\section{Introduction}
The traditional supervised end-to-end training~\cite{DBLP:journals/corr/BojarskiTDFFGJM16} of any deep learning system aims at training the model to perform specific task by providing supervision in the form of pairs of input data along with their corresponding desired outputs. In this framework model by itself creates its own internal representation of the problem and decides which features of the input data are relevant to the problem. This approach has a crucial drawback. In particular, one has no direct influence on the process of constructing the internal data representation or the selection of relevant features of the input that the model should focus on when learning. This may lead to creating wrong correlations between the input and output, i.e., model may potentially form prediction based on features which in general are not relevant to the problem, but are nevertheless present in the training data. Moreover, debugging such system is not straightforward and requires dedicated tools such as visualization techniques which identify features of the input as well intermediate network layers that are selected by the model to form the prediction. 

We propose to leverage visualization techniques, in particular VisualBackProp~\cite{DBLP:journals/corr/BojarskiCCFJMZ16} , to guide the model during the training process to look for only the relevant features in the data. This provides an additional supervision to the learning system. The second supervision can be seen as the privileged information available during training (but \textit{not} available during testing) and therefore, the proposed double supervision framework fits well into the Learning with Privileged Information (LUPI) paradigm~\cite{DBLP:journals/nn/VapnikV09}. The LUPI paradigm was explored before in the context of support vector machines (SVMs). We show a technique that inserts the privilege information during training of deep learning models. The enabling mechanism here relies on two characteristic aspects of VisualBackProp: i) the method shows which parts of the input image contribute most to the prediction of the network and ii) the method uses only linear operations (averaging, deconvolution, and point-wise multiplication) typical for neural networks and thus can be built-in to any CNN as an additional architectural component of the model. In case of the image classification problem, the segmentation mask can be treated as a privileged information, i.e. an extra supervision provided additionally to the image label. In this case, the latter property of VisualBackProp automatically enables back-propagating the desired visualization mask for the input image and focuses networks attention on concrete sub-region of the image when forming a prediction. 

The paper is organized as follows: Section~\ref{sec:rw} provides literature overview, Section~\ref{sec:pa} discusses how to use VisualBackProp for learning using privileged information with CNNs, Section~\ref{sec:Exp} provides extensive empirical evaluation of the approach, and finally Section~\ref{sec:Conclusions} summarizes the work.

\section{Related Work}
\label{sec:rw}
LUPI framework was theoretically studied in the literature~\cite{DBLP:conf/nips/PechyonyV10} and applied almost exclusively in the settings involving SVMs~\cite{JMLR:v16:vapnik15b,DBLP:journals/nn/LapinHS14,Feyereisl:2014:OLB:2968826.2968850,DBLP:conf/iccv/SharmanskaQL13,Ribeiro:2012:EDR:2181339.2181697,4634079}
. Recently, LUPI was employed in the deep learning setting~\cite{DBLP:conf/ijcai/ChenJFY17}, where the authors propose a novel group orthogonal convolutional neural network (GoCNN) that learns untangled representations within each layer by utilizing the privileged information in the form of a segmentation mask. In particular, the feature maps of the last convolutional layer in GoCNN are divided into two parts corresponding to the foreground and background respectively. The low-resolution segmentation mask (the resolution is dictated by the last feature maps) is used to suppress either the background or foreground. The additional fully-connected layers utilized only during training (and removed at testing) learn the classification label for the image based on only the foreground or only the background, thus they encourage the network to learn relevant features in two mentioned groups. The final prediction is done based on the joint feature representation of the image foreground and background. As opposed to this approach we instead utilize the visualization technique to take the privileged information into account at training. We use the segmentation mask at the full resolution of the image and suppress irrelevant information in all convolutional layers of the network. We use the same architecture at training and testing. The enabling mechanism relies on the fact that the visualization technique we use can be built-in to any convolutional network, is parameter-free, and requires order of magnitude less computations than the forward pass.

We next provide a brief overview of the existing visualization techniques. They aim at understanding which parts of the input data most contribute to the prediction of the network. These works are typically focusing on convolutional neural networks (CNNs) and consider image data. A notable approach~\cite{BachPLOS15} proposes a methodology called layer-wise relevance propagation (LRP), where the relevance of each neuron is redistributed to its predecessors through a particular message-passing scheme relying on the conservation principle. LRP was later extensively studied and extended~\cite{BinBacMonMueSam16,bach-arxiv15,DBLP:journals/corr/ArrasHMMS16}. An extensive comparison of LRP with other techniques~\cite{DBLP:journals/corr/SamekBMBM15}, like the deconvolution method~\cite{MR14} (deconvolutional method uses deconvolutional neural network~\cite{MGR11} and was analyzed in~\cite{DBLP:journals/corr/SimonyanVZ13,BachPLOS15}) and the sensitivity-based approach~\cite{DBLP:journals/corr/SimonyanVZ13,Baehrens:2010:EIC:1756006.1859912,Rasmussen_visualizationof} reveals that LRP provides better explanation of the CNN classification decisions than considered competitors. VisualBackProp (we refer the reader to this work~\cite{DBLP:journals/corr/BojarskiCCFJMZ16} for extensive review of all existing visualization techniques) that we rely on in this paper provides very similar performance to LRP and requires order of magnitude less computations. To the best of our knowledge, we are the first to incorporate the visualization technique into the training process of the deep learning model. Applying the visualization methods in the LUPI framework in general is challenging as any such method has to fit well to the network training process, has to allow back-propagation, and preferably needs to be computationally efficient.

\section{Proposed Approach}
\label{sec:pa}
In this section, we briefly review the LUPI framework and then we describe our approach for incorporating the privileged information in the training of deep convolutional networks via VisualBackProp.
\subsection{LUPI framework}
In a classical training paradigm, we are given $n$ training examples with their corresponding labels, i.e. $(x_{1},y_{1}),(x_{2},y_{2}), \dots, (x_{n},y_{n})$ , obtained from an unknown joint probability distribution $P(x,y)$. The objective is to find a function $f(x,\alpha^{*})$ chosen from a hypothesis space $\mathcal{F} = \{ f(x;\alpha): \alpha \in \Omega\}$ consisting of classification rules parameterized by $\alpha$ ($\Omega$ denotes the parameter space) that minimizes the risk of incorrect classification. The problem can be formulated as a minimization problem of the form
\begin{equation}
    R(\alpha) = \frac{1}{2} \int_{}{}L(y,f(x;\alpha))d P(x,y),
    \label{eq:risk}
\end{equation}
where $L(y,f(x,\alpha))$ is the classification loss. In the LUPI framework, at training the learning machine is provided with $(x,y)$ pairs augmented by an additional information $x^{*}$ that is referred to as privileged information and is not available at testing. Available training samples $(x_{1},x^{*}_{1},y_{1}), (x_{2},x^{*}_{2},y_{2}), \dots, (x_{n},x^{*}_{n},y_{n})$ are generated from a fixed but unknown distribution $P(x,x^{*},y)$. The  privileged information $x^{*}$ is available in many learning problems. For example, in the context of image classification, such information is contained in segmentation masks, object bounding boxes, object contours, etc. The learning problem changes to
\begin{equation}
    R^{'}(\alpha) = \frac{1}{2} \int_{}{}L(y,f_{x^{*}}(x;\alpha))d P(x,x^{*},y),
    \label{eq:risk2}
\end{equation}
where $f_{x^{*}}: \mathcal{X} \rightarrow \mathbb{R}$ does not rely on the privileged data $x{*}$, since this data is not available at testing, however this data guide the choice of hypothesis $f_{x^{*}}$ from $\mathcal{F}$. Incorporating the privileged information is expected to improve the generalization of the resulting model.

\subsection{VisualBackProp}
The VisualBackProp visualization technique is based on the intuition that when moving from the input to the output of the CNN, the feature maps contain less irrelevant information but at the same time the deeper feature maps have lower resolution. The method obtains the visualization mask by``projecting''  the relevant information of the last convolutional layer of the network to its higher-resolution input. VisualBackProp utilizes the forward propagation pass, which is already done to obtain a prediction and thus do not require any additional forward passes. The method uses the thresholded feature maps obtained after each ReLU layer. In the first step, the feature maps from each layer are averaged, resulting in a single feature map per layer. Next, the averaged feature map of the deepest convolutional layer is scaled up to the size of the feature map of the previous layer. This is done using deconvolution with filter size and stride that are the same as the ones used in the deepest convolutional layer (for deconvolution the same filter size and stride are used as in the convolutional layer which outputs the feature map that we are scaling up with the deconvolution). In deconvolution, all weights are set to $1$ and biases to $0$. The obtained scaled-up averaged feature map is then point-wise multiplied by the averaged feature map from the previous layer. The resulting image is again scaled via deconvolution and multiplied by the averaged feature map of the previous layer exactly as described above. This process continues all the way to the network’s input. The resulting mask is normalized to the range $[0, 1]$. In order to handle complex architectures, i.e. ResNets, we treat the whole ResNet block as single convolutional layer with non-linearity. Thus, in the VisualBackProp, we use single deconvolutional layer for each ResNet block and we use feature maps on the output of the corresponding block. 

\subsection{VisualBackProp for LUPI framework}

\begin{figure}[h!]
\centerline{\includegraphics[width=1.0\textwidth]{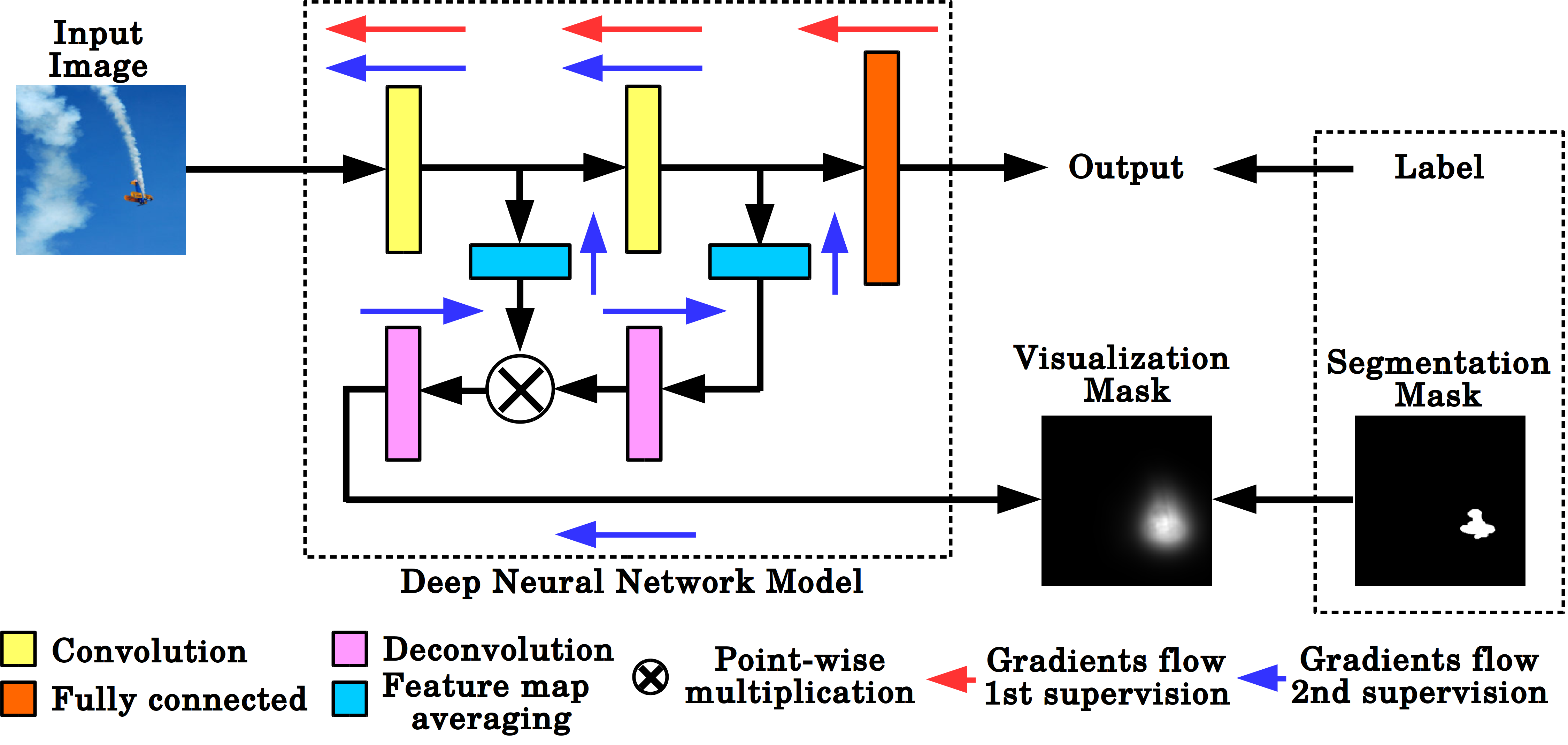}}
\caption{Block diagram of the proposed VisualBackProp + LUPI training.}
\label{fig:dualSupervision}
\vspace*{-0.05in}
\end{figure}

In the LUPI framework the privileged information is only available at training, which makes it challenging to incorporate it into the classical learning settings, among them deep learning. As the privileged information is not available at test time, it can not be used at the input of the neural network. In this work, we propose to utilize the VisualBackProp visualization technique to incorporate the privileged information at training. Recall that VisualBackProp method uses operations such as averaging, deconvolution, and point-wise multiplication, which can be implemented as the neural network layers. Thus, VisualBackProp can be built into any CNN architecture. As VisualBackProp becomes an integral part of the neural network architecture, it allows for backward propagation through the output corresponding to visualization mask. Thus, one can put constraints on the visualization mask during the training phase. In particular, one can train the neural network to output the desired visualization mask for the particular input, which is equivalent to guiding the neural network to take into account only the desired parts of the input when forming the prediction. The training with such double supervision  is illustrated in Figure~\ref{fig:dualSupervision}.

Incorporating the visualization architecture into the deep learning system is more effective than keeping them separate and alternate between them when training since the visualization mask can be obtained during the forward pass together with the original neural network output. Furthermore, this solution allows for better utilization of resources (i.e., GPU utilization). 

When training the network, our approach uses an additional term in the cost function, which utilizes the privileged information in the form of the segmentation mask to enforce the network to focus on the relevant parts of the input image. The proposed cost term $L_{PI}$ that incorporates the privileged information in training is based on the difference on the visualization mask $I_{vis}(x;\alpha)$ obtained from VisualBackProp and segmentation mask $I_{seg}(x)$ that correspond to image $x$. Note that $I_{seg}(x)$ constitutes the privileged information thus $x^{*} = I_{seg}(x)$. The visualization mask is updated during training as a result of the updates of network parameters $\alpha$. We consider two variants of the proposed cost term:

\vspace{-0.05in}
\begin{enumerate}
\item \textit{Full focus}: where we encourage the network to observe the entire segmented region and penalize when it is looking outside the segmented region:

\vspace{-0.1in}
\begin{equation}
    L_{PI}(x;\alpha) = || I_{vis}(x;\alpha) - I_{seg}(x) ||_{1}
    \label{eq:full_focus}
\end{equation}

\item \textit{Half focus}: where we only penalize the network if it looks outside the segmented region, but we do not force the network to look at the entire segmented region (thus the network can rely only on the part of this region in practice when forming a prediction)
\begin{equation}
    L_{PI}(x;\alpha) = || I_{vis}(x;\alpha) - I_{vis}(x;\alpha) \odot I_{seg}(x) ||_{1},
    \label{eq:half_focus}
\end{equation}
where $\odot$ is an element wise multiplication. 
\end{enumerate}

The additional loss term $L_{PI}(x;\alpha)$ is added to the classification loss, i.e. cross-entropy as shown in Equation~\ref{eq:totloss}. 
\begin{equation}
    L(x;\alpha) = -y\log f_{x^{*}}(x;\alpha) - (1-y)\log(1-f_{x^{*}}(x;\alpha)) + \lambda*L_{PI}(x;\alpha).
    \label{eq:totloss}
\end{equation}
The proposed methodology leverages the visualization technique to guide the model during the training process to look for relevant features in the data. This provides an additional supervision to the learning system and to some extend counteracts the ``black-box'' effect characteristic of the end-to-end deep learning models.

At testing, the additional architectural component of the CNN model that is induced by VisualBackProp operations generates a visualization mask for the input image, but do not contribute to the prediction of the image label made by the network.

\section{Experiments}
\label{sec:Exp}
To demonstrate the effectiveness of our training paradigm we show the improvements in the evaluation metrics on multiple data sets: ranging from multi-label multi-instance image detection data sets such as PASCAL VOC~\cite{Everingham10} and ImageNet~\cite{ILSVRC15} to ISIC skin lesion data set~\cite{ISIC} which consists of dermoscopic images collected with high-resolution digital dermatoscopes.
\subsection{PASCAL VOC}
\begin{wraptable}{l}{0.55\textwidth}
\vspace{-0.15in}
    \centering
    \caption{Average Precision in each class obtained on the PASCAL VOC test data set.}
    \begin{tabular}{|c |c |c |c|}
         \hline
         Category &     Regular & Half focus & Full focus  \\
         \hline
        \hline
         aeroplane	 &  0.986 & \textbf{0.987}    & 0.980 \\
         bicycle &  	0.881 & \textbf{0.895}    & 0.876 \\
         bird &     	0.944 & \textbf{0.950}    & 0.932 \\
         boat &     	0.916 & \textbf{0.923}    & 0.887 \\
         bottle &   	0.674 & \textbf{0.700}    & 0.657 \\
         bus &  	    0.900 & \textbf{0.915}    & \textbf{0.903} \\
         car &  	    0.835 & \textbf{0.861}    & \textbf{0.855} \\
         cat &  	    0.969 & \textbf{0.970}    & 0.957 \\
         chair &    	0.734 & \textbf{0.782}    & \textbf{0.766} \\
         cow &  	    0.814 & \textbf{0.904}    & \textbf{0.846} \\
         diningtable &  0.798 & \textbf{0.803}    & 0.790 \\
         dog &  	    0.955 & \textbf{0.958}    & 0.937 \\
         horse &    	0.920 & \textbf{0.953}    & \textbf{0.921} \\
         motorbike &    0.931 & \textbf{0.935}    & 0.921 \\
         person &   	0.961 & \textbf{0.966}    & \textbf{0.962} \\
         pottedplant &  0.543 & \textbf{0.652}    & \textbf{0.626} \\
         sheep &    	0.864 & \textbf{0.909}   & \textbf{0.871} \\
         sofa &     	0.710 & \textbf{0.750}    & \textbf{0.748} \\
         train &    	0.956 & \textbf{0.963}    & 0.952 \\
         tvmonitor &    0.862 & \textbf{0.874}    & \textbf{0.864} \\
         \hline
        \hline
         Mean & 0.858 & \textbf{0.882} & \textbf{0.863} \\
         \hline
    \end{tabular}
    \label{tab:pascal}
\vspace{-0.1in}
\end{wraptable}

\begin{figure}[b!]
\vspace{-0.25in}
    \centering
    \begin{minipage}{\textwidth}
    \includegraphics[width=0.12\textwidth]{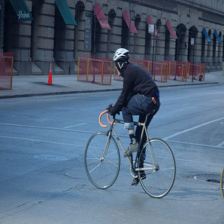}
    \includegraphics[width=0.12\textwidth]{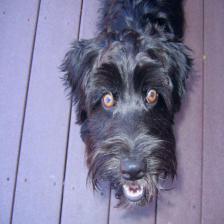}
    \includegraphics[width=0.12\textwidth]{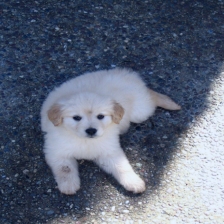}
    \includegraphics[width=0.12\textwidth]{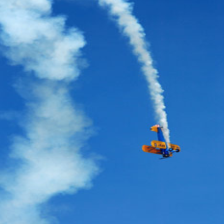}
    \includegraphics[width=0.12\textwidth]{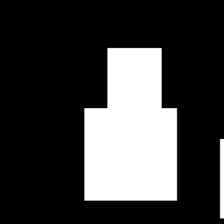}
    \includegraphics[width=0.12\textwidth]{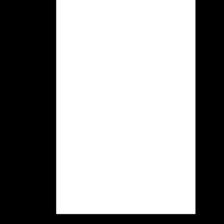}
    \includegraphics[width=0.12\textwidth]{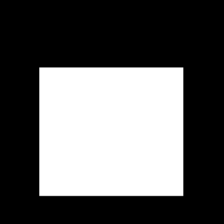}
    \includegraphics[width=0.12\textwidth]{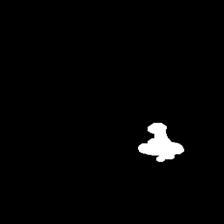}
    \end{minipage}

    \begin{minipage}{\textwidth}
    \includegraphics[width=0.12\textwidth]{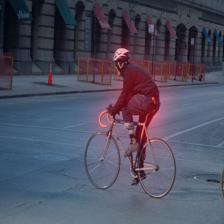}
    \includegraphics[width=0.12\textwidth]{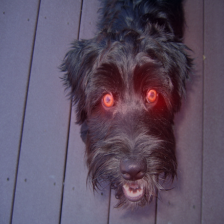}
    \includegraphics[width=0.12\textwidth]{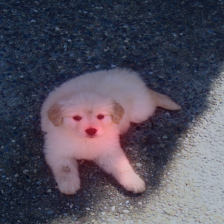}
    \includegraphics[width=0.12\textwidth]{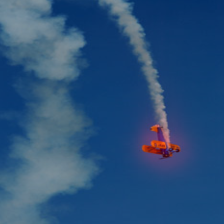}
    \includegraphics[width=0.12\textwidth]{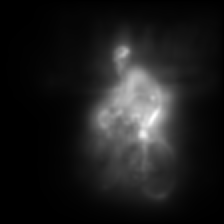}
    \includegraphics[width=0.12\textwidth]{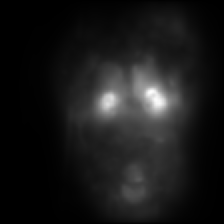}
    \includegraphics[width=0.12\textwidth]{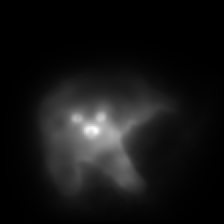}
    \includegraphics[width=0.12\textwidth]{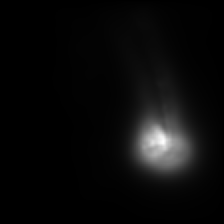}
    \end{minipage}

    \begin{minipage}{\textwidth}
    \includegraphics[width=0.12\textwidth]{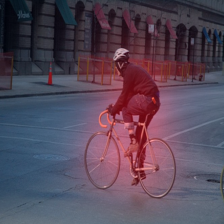}
    \includegraphics[width=0.12\textwidth]{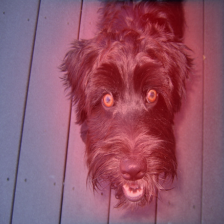}
    \includegraphics[width=0.12\textwidth]{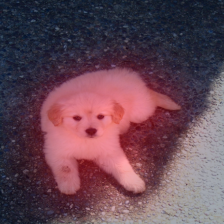}
    \includegraphics[width=0.12\textwidth]{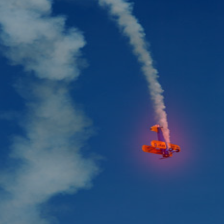}
    \includegraphics[width=0.12\textwidth]{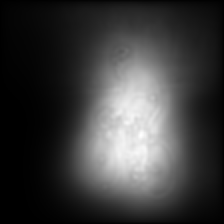}
    \includegraphics[width=0.12\textwidth]{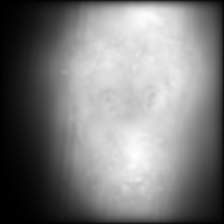}
    \includegraphics[width=0.12\textwidth]{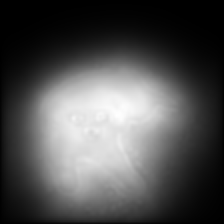}
    \includegraphics[width=0.12\textwidth]{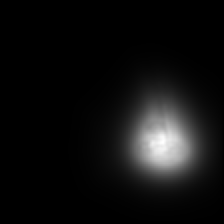}
    \end{minipage}
    
    \begin{minipage}{\textwidth}
    \includegraphics[width=0.12\textwidth]{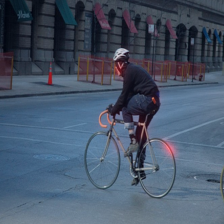}
    \includegraphics[width=0.12\textwidth]{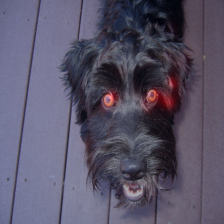}
    \includegraphics[width=0.12\textwidth]{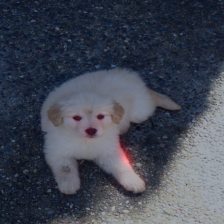}
    \includegraphics[width=0.12\textwidth]{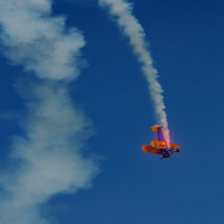}
    \includegraphics[width=0.12\textwidth]{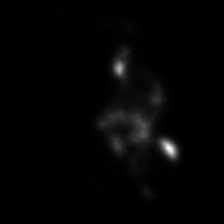}
    \includegraphics[width=0.12\textwidth]{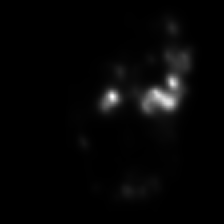}
    \includegraphics[width=0.12\textwidth]{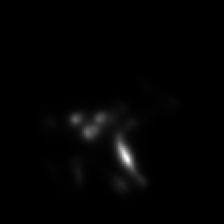}
    \includegraphics[width=0.12\textwidth]{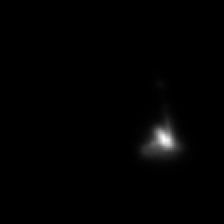}
    \end{minipage}

    \caption{\label{fig:pascal_images} PASCAL VOC. \textbf{First row}: Exemplary input images with their corresponding segmentation masks. \textbf{Following rows}: visualization masks (\textbf{right four images}) and the mask overlaid in red on the input image (\textbf{left four images}) obtained with Regular training \textbf{(second row)}, Full focus (\textbf{third row}), and Half focus (\textbf{fourth row}).}
    \vspace{-0.2in}
\end{figure}
The detection task in the 2014 PASCAL VOC challenge is a multi-label classification problem with $20$ classes. The data set does not contain segmentation masks for all the images. The challenge uses average precision in each class on the test data to rank submissions. In this experiment we used ResNet-50 model, where the final fully connected layer was replaced to output scores over $20$ classes. We used sigmoid function on the network's output. The binary cross-entropy criterion was used to train the model. We ran three separate experiments: i) utilizing only binary cross-entropy loss, ii) full focus scenario based on Equation~\ref{eq:full_focus}, and iii) half focus scenario based on Equation~\ref{eq:half_focus}. The Adam optimizer with a learning rate of $1e^{-4}$ and beta values $0.9$ and $0.99$ was employed during initial rounds of training. Once the model starts over-fitting we switch to stochastic gradient descent (SGD) with a learning rate of $1e^{-4}$ and momentum value of $0.9$. The hyper-parameters were kept fixed between the experiments. The submissions were made through the PASCAL VOC submission portal. Results are shown in Table~\ref{tab:pascal} and reveal the superiority of introducing the privileged information to deep model training via VisualBackProp over regularly supervised training. Finally, Figure~\ref{fig:pascal_images} shows visualization masks for different methods on a few exemplary test images. In case of Full focus and Half focus methods the visualization architecture was used at training providing an additional supervision. In Half focus case the visualization mask is much sharper than in any other method. Full focus method is able to well-capture the entire region of interest dictated by the visualization mask when forming a prediction.

\subsection{ImageNet}
\begin{wraptable}{l}{0.55\textwidth}
\vspace{-0.19in}
    \centering
    \caption{Average Precision in each class obtained on the ImageNet detection challenge validation data-set.}
\vspace{-0.05in}
    \begin{tabular}{|c| c| c| c|}
    \hline
    Category	& Regular & Half focus & Full focus \\
    \hline
    \hline
    accordion	& \textbf{0.576}           & 0.568 &  0.573\\
    airplane	& 0.470           & \textbf{0.496} &  0.462\\
    ant		    & 0.443           & \textbf{0.465} &  \textbf{0.474}\\
    antelope	& 0.327           & \textbf{0.346} &  0.326\\
    apple		& 0.136           & \textbf{0.178} &  0.129\\
    
    \multicolumn{4}{c}{... (190 categories)} \\      
    water bottle	& 0.102       & \textbf{0.130} &  0.083\\
    watercraft	& 0.447           & \textbf{0.468} &  0.429\\
    whale		& 0.605           & \textbf{0.638} &  0.591\\
    wine bottle	& 0.185           & \textbf{0.214} &  0.182\\
    zebra		& \textbf{0.377}           & 0.368 &  0.372\\
    \hline 
\hline
    Mean:	 & 0.260          & \textbf{0.287} &  0.248\\
    \hline 
    Median:	 & 0.219          & \textbf{0.243} &  0.198\\
    \hline
    \end{tabular}

    \label{tab:imagenet_results}
\vspace{-0.15in}
\end{wraptable}

The ImageNet detection challenge  is a multi-label classification problem with $200$ categories and over $456,567$ training images with over $534,309$ object instances. Similar to PASCAL VOC experiment we replaced the final fully connected layer to output probabilities over $200$ object categories.

\begin{figure}[htp!]
    \centering
    \begin{minipage}{\textwidth}
    \includegraphics[width=0.12\textwidth]{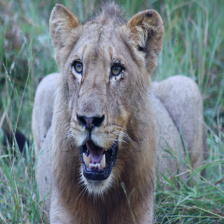}
    \includegraphics[width=0.12\textwidth]{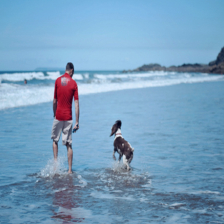}
    \includegraphics[width=0.12\textwidth]{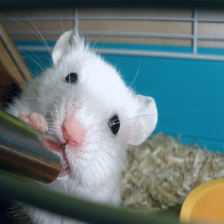}
    \includegraphics[width=0.12\textwidth]{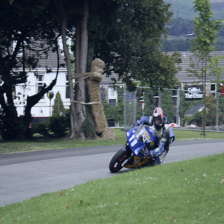}
    \includegraphics[width=0.12\textwidth]{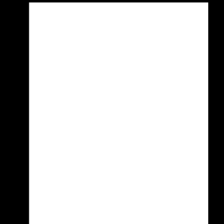}
    \includegraphics[width=0.12\textwidth]{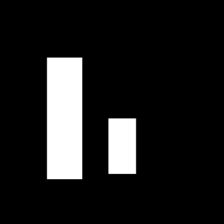}
    \includegraphics[width=0.12\textwidth]{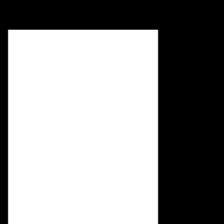}
    \includegraphics[width=0.12\textwidth]{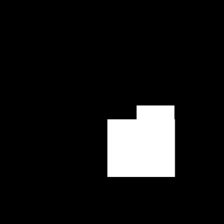}
    \end{minipage}
    
    \begin{minipage}{\textwidth}
    \includegraphics[width=0.12\textwidth]{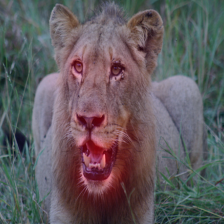}
    \includegraphics[width=0.12\textwidth]{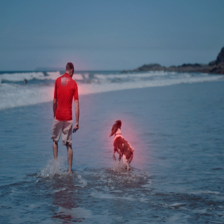}
    \includegraphics[width=0.12\textwidth]{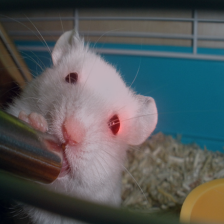}
    \includegraphics[width=0.12\textwidth]{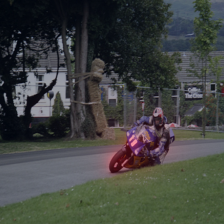}
    \includegraphics[width=0.12\textwidth]{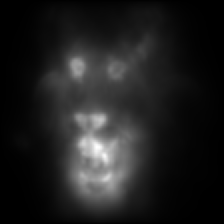}
    \includegraphics[width=0.12\textwidth]{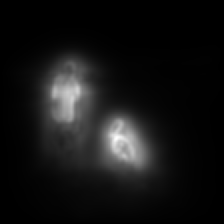}
    \includegraphics[width=0.12\textwidth]{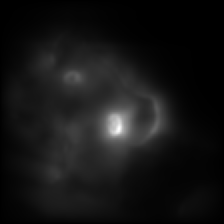}
    \includegraphics[width=0.12\textwidth]{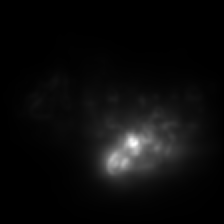}
    \end{minipage}
    
    \begin{minipage}{\textwidth}
    \includegraphics[width=0.12\textwidth]{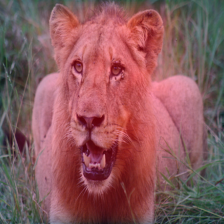}
    \includegraphics[width=0.12\textwidth]{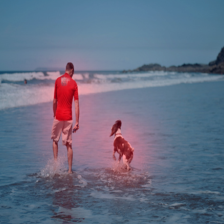}
    \includegraphics[width=0.12\textwidth]{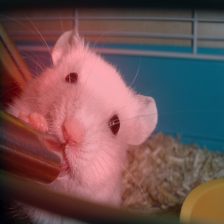}
    \includegraphics[width=0.12\textwidth]{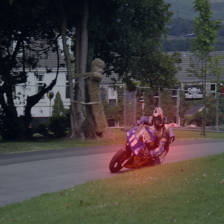}
    \includegraphics[width=0.12\textwidth]{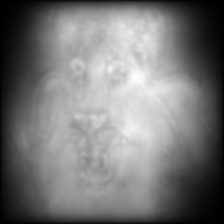}
    \includegraphics[width=0.12\textwidth]{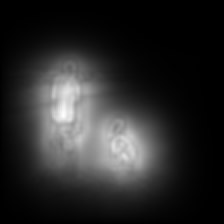}
    \includegraphics[width=0.12\textwidth]{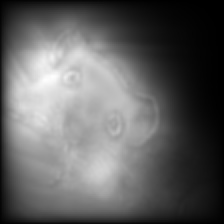}
    \includegraphics[width=0.12\textwidth]{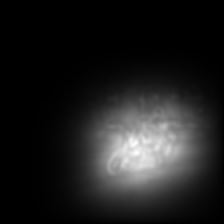}
    \end{minipage}
    
    \begin{minipage}{\textwidth}
    \includegraphics[width=0.12\textwidth]{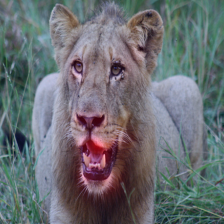}
    \includegraphics[width=0.12\textwidth]{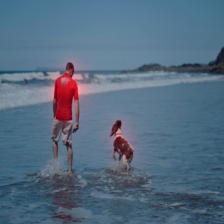}
    \includegraphics[width=0.12\textwidth]{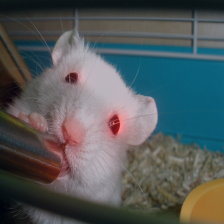}
    \includegraphics[width=0.12\textwidth]{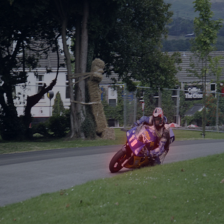}
    \includegraphics[width=0.12\textwidth]{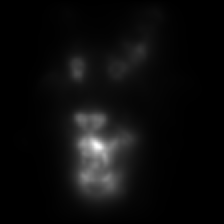}
    \includegraphics[width=0.12\textwidth]{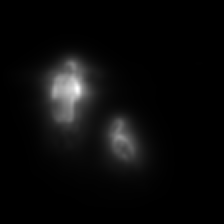}
    \includegraphics[width=0.12\textwidth]{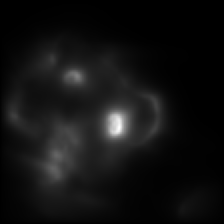}
    \includegraphics[width=0.12\textwidth]{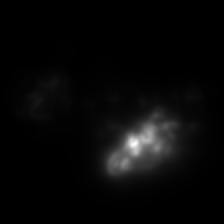}
    \end{minipage}

    \caption{\label{fig:imagenet_images} ImageNet. \textbf{First row}: Exemplary input images with their corresponding segmentation masks. \textbf{ The following rows}: visualization masks (\textbf{right four images}) and the visualization masks in red overlaid on the input image (\textbf{left four images}) obtained with Regular training \textbf{(second row)}, Full focus (\textbf{third row}), and Half focus (\textbf{fourth row}).}
\end{figure}

We train the ResNet-50 model with Adam optimizer and a learning rate of $1e-4$ and later switch to SGD with learning rate of $4e-3$. This learning rate was decayed gradually to $1e-4$. The results shown in Table~\ref{tab:imagenet_results} are obtained on the ILSVRC2013 validation data-set. None of the images in this data set is paired with a corresponding segmentation mask, but instead there are bounding boxes provided for each object from the $200$ categories in every image. We treat those boxes as segmentation masks. The performance quality of Full focus method is affected by the low quality of the resulting segmentation masks and was worse than the regular training of the network. Note however that the Half focus method is robust to imprecise segmentation masks due to its nature, i.e. the method prevents the model to look outside the mask but within the area of the mask it can decide on the most important regions for forming the prediction as opposed to the Full focus setting where the model is forced to look at the entire area of the segmentation mask when forming the prediction. It is also intuitive to understand why this is happening since just as the high quality privileged information promotes better generalization of the model, low quality one can result in performance deterioration. Similarly to Figure~\ref{fig:pascal_images}, Figure~\ref{fig:imagenet_images} shows visualization masks for different methods on a few exemplary test images.

\begin{figure}[htp!]
\vspace{-0.1in}
\centering
    \includegraphics[width=0.34\textwidth]{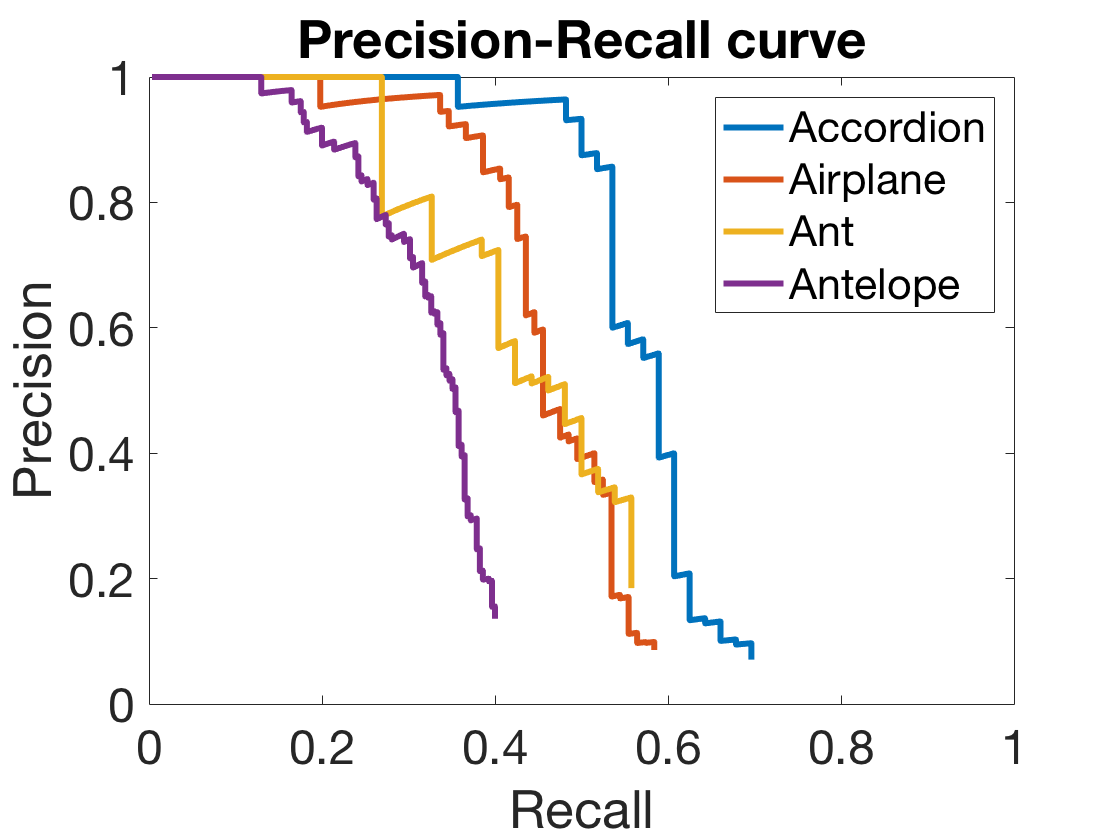}
    \hspace{-0.1in}\includegraphics[width=0.34\textwidth]{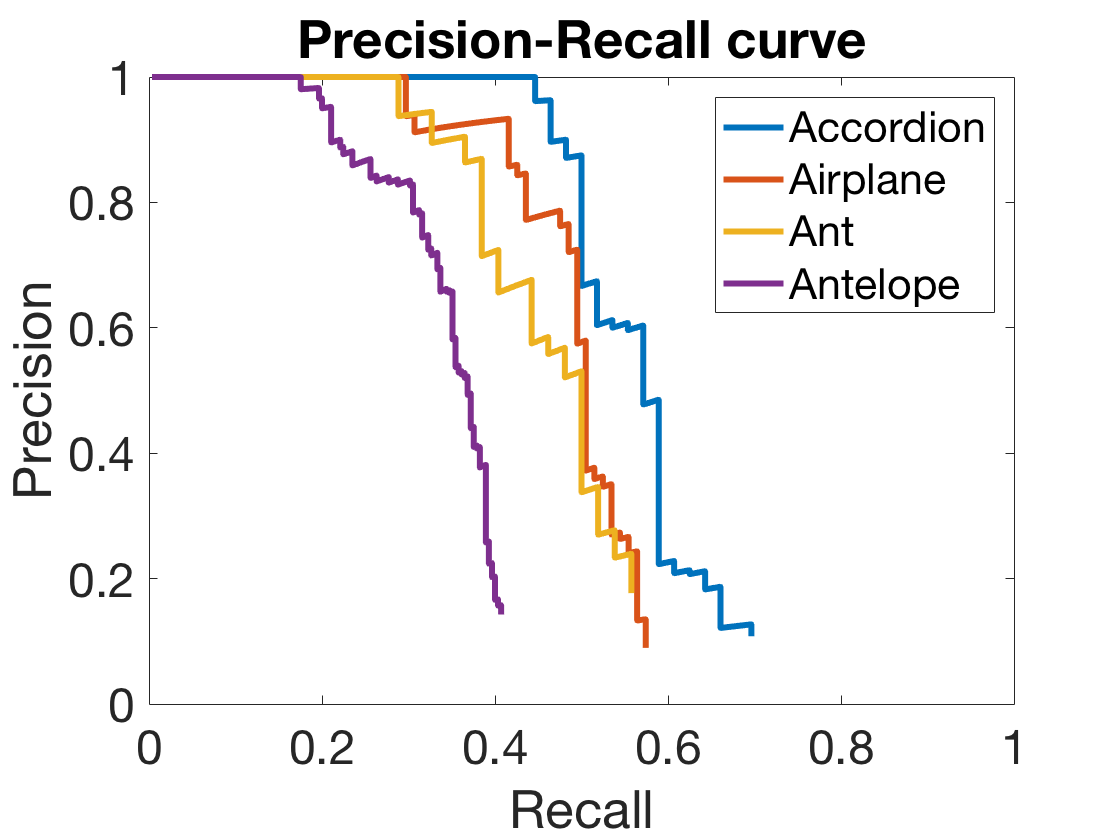}
    \hspace{-0.1in}\includegraphics[width=0.34\textwidth]{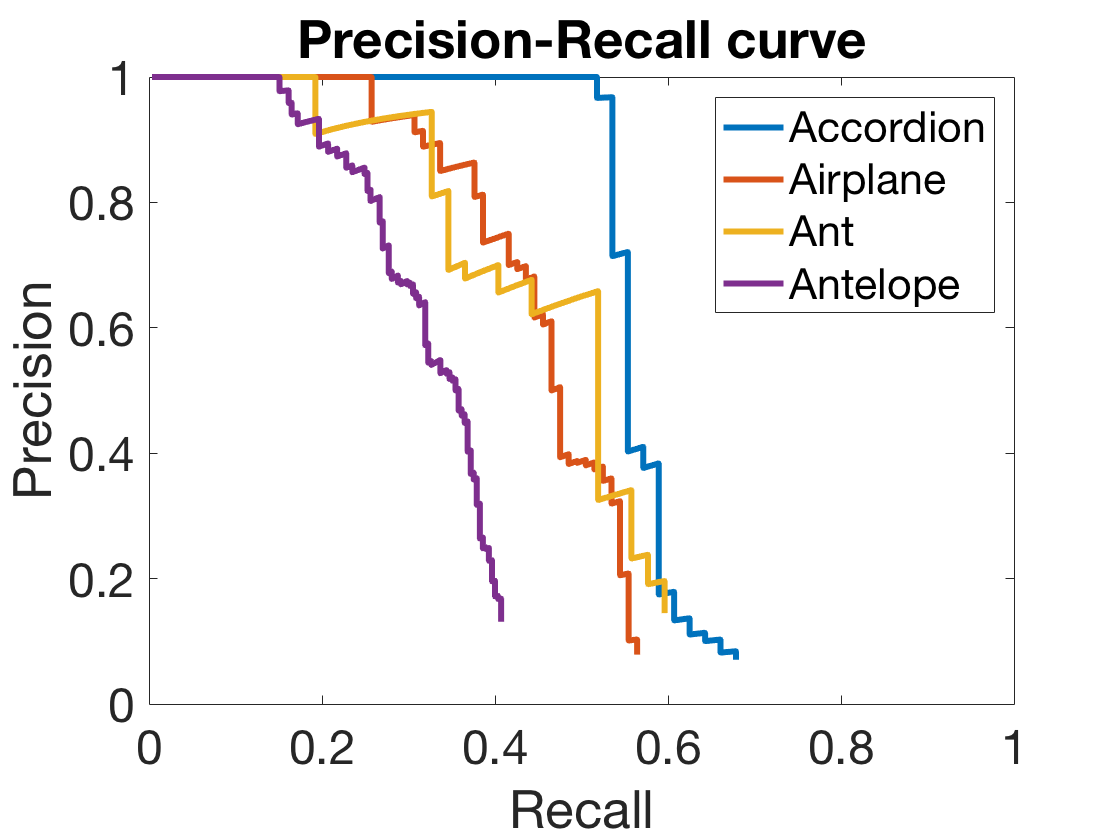}
    \caption{Precision-Recall Curves for $4$ exemplary classes (accordion, airplane, ant, and antelope) on ImageNet dataset. \textbf{From left to right:} Regular training, Half Focus, and Full Focus }
    \label{fig:my_label2}
\end{figure}

Figure~\ref{fig:my_label2} depicts precision-recall curves for $4$ randomly chosen classes: accordion, airplane, ant, and antelope. 

\subsection{Melanoma Classification}

\begin{table}[H]
\vspace{-0.1in}
\caption{Receiver operating characteristics metrics for skin lesion classification.}
\begin{tabular}{|L{2.5cm}|c|c|c|c|}
    \hline
    & \multicolumn{4}{c|}{ROC-AUC} \\
    \cline{2-5}
    Experiment &  Melanoma & Seborrheic Keratosis & Micro-Average& Macro-Average\\
    &  (ROC AUC) & (ROC AUC) &  ROC&  ROC\\
    \hline
    \hline
    Regular & 0.802 & 0.924 & 0.904 & 0.879 \\
    \hline
    Added seg. mask& 0.729 & 0.904 & 0.867 & 0.827 \\
    \hline
    Seg. mole& 0.726 & 0.895 & 0.847 & 0.818 \\
    \hline
    Full focus & \textbf{0.815} & 0.939 & \textbf{0.912} & \textbf{0.883} \\
    \hline
    Half focus  & 0.807 & \textbf{0.941} & 0.911 & 0.881 \\
    \hline
\end{tabular}
\label{tab:melanoma}
\end{table}

\begin{figure}[H]
\vspace{-0.3in}
    \centering
    \includegraphics[width=0.32\textwidth]{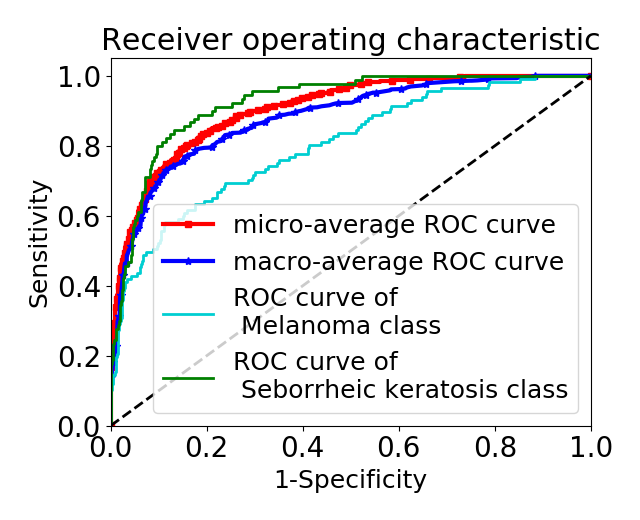}
    \includegraphics[width=0.32\textwidth]{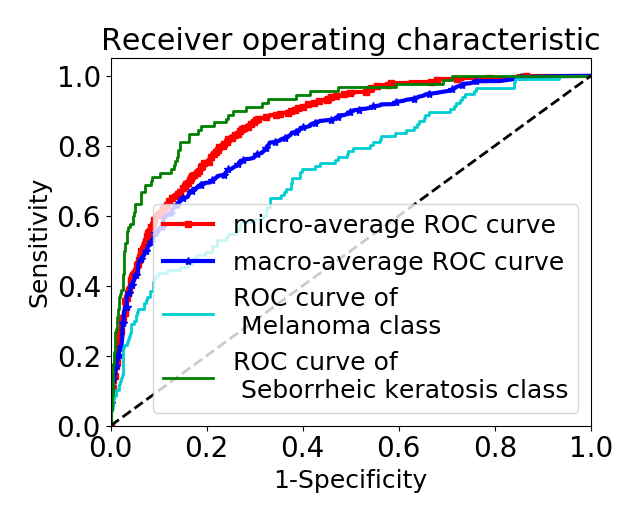}
    \includegraphics[width=0.32\textwidth]{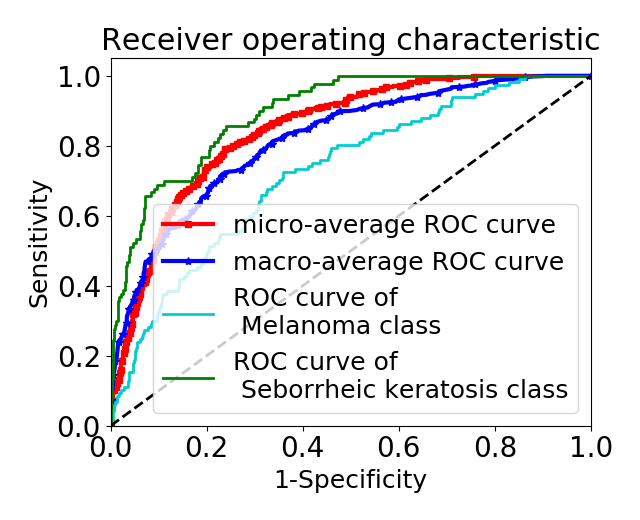}\\
\vspace{-0.05in}
    \includegraphics[width=0.32\textwidth]{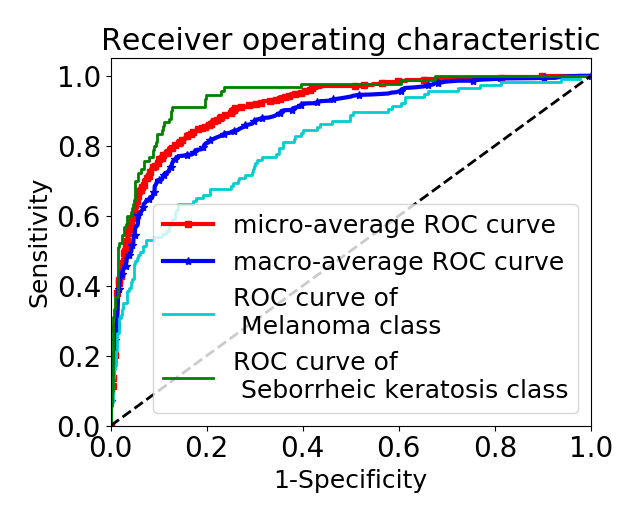}
    \includegraphics[width=0.32\textwidth]{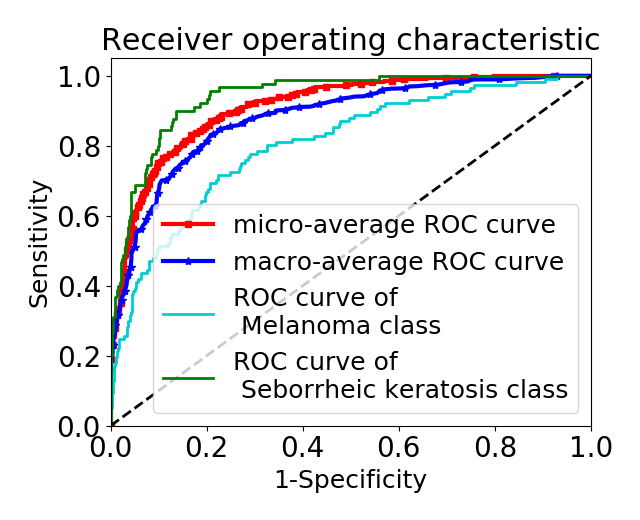}
\vspace{-0.08in}
    \caption{\label{fig:roc_curves} ROC curves for different methods. \textbf{First row}: Regular training (\textbf{left}), Added seg. mask (\textbf{middle}), Seg. mole (\textbf{right}), \textbf{Second row}: Full focus (\textbf{left}) and Half focus (\textbf{right}).}
\vspace{-0.17in}
\end{figure}

\begin{figure}[H]
\vspace{-0.15in}
\centering
\begin{minipage}{\textwidth}
\includegraphics[width=0.12\textwidth]{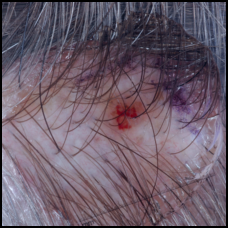}
\includegraphics[width=0.12\textwidth]{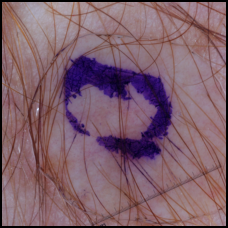}
\includegraphics[width=0.12\textwidth]{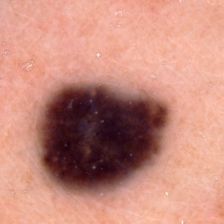}
\includegraphics[width=0.12\textwidth]{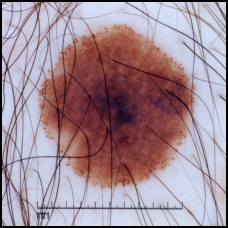}
\includegraphics[width=0.12\textwidth]{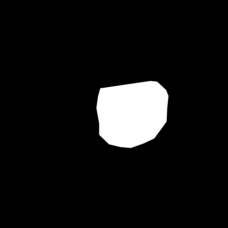}
\includegraphics[width=0.12\textwidth]{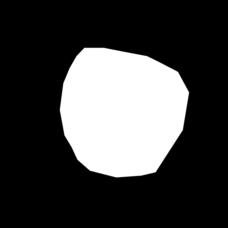}
\includegraphics[width=0.12\textwidth]{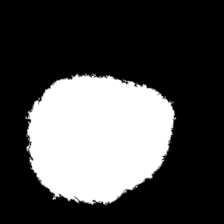}
\includegraphics[width=0.12\textwidth]{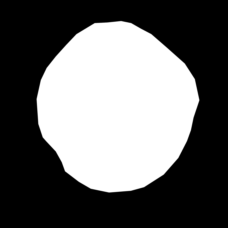}
\end{minipage}

\begin{minipage}{\textwidth}
\includegraphics[width=0.12\textwidth]{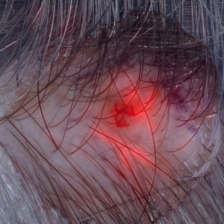}
\includegraphics[width=0.12\textwidth]{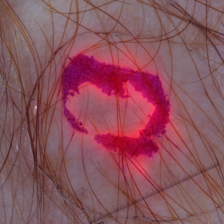}
\includegraphics[width=0.12\textwidth]{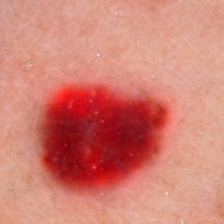}
\includegraphics[width=0.12\textwidth]{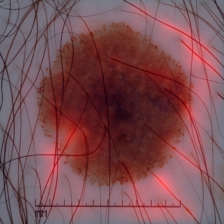}
\includegraphics[width=0.12\textwidth]{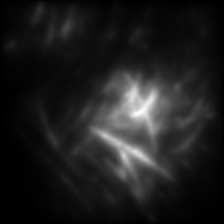}
\includegraphics[width=0.12\textwidth]{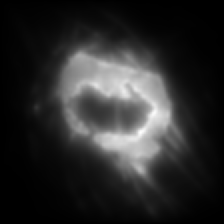}
\includegraphics[width=0.12\textwidth]{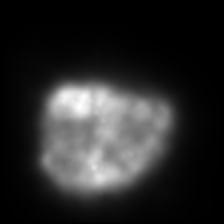}
\includegraphics[width=0.12\textwidth]{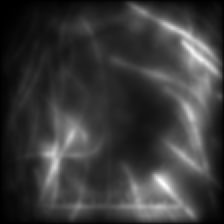}
\end{minipage}

\begin{minipage}{\textwidth}
\includegraphics[width=0.12\textwidth]{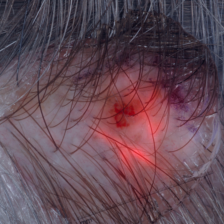}
\includegraphics[width=0.12\textwidth]{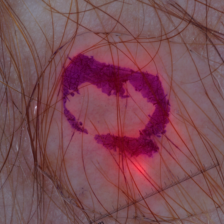}
\includegraphics[width=0.12\textwidth]{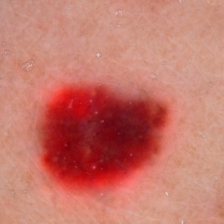}
\includegraphics[width=0.12\textwidth]{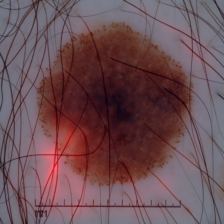}
\includegraphics[width=0.12\textwidth]{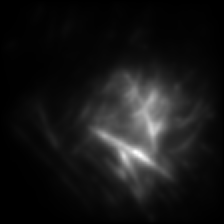}
\includegraphics[width=0.12\textwidth]{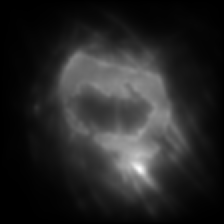}
\includegraphics[width=0.12\textwidth]{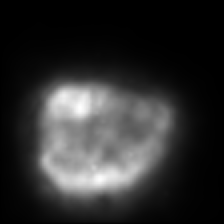}
\includegraphics[width=0.12\textwidth]{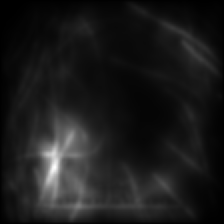}
\end{minipage}

\begin{minipage}{\textwidth}
\includegraphics[width=0.12\textwidth]{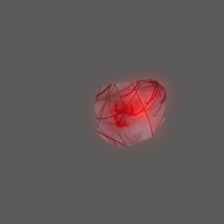}
\includegraphics[width=0.12\textwidth]{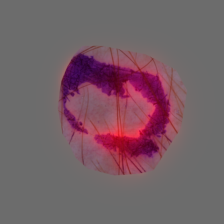}
\includegraphics[width=0.12\textwidth]{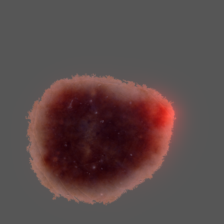}
\includegraphics[width=0.12\textwidth]{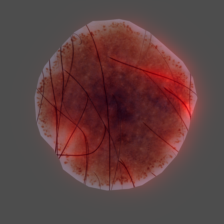}
\includegraphics[width=0.12\textwidth]{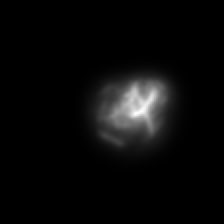}
\includegraphics[width=0.12\textwidth]{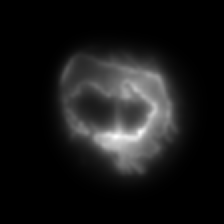}
\includegraphics[width=0.12\textwidth]{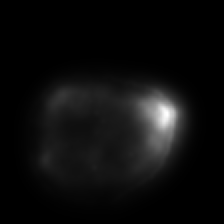}
\includegraphics[width=0.12\textwidth]{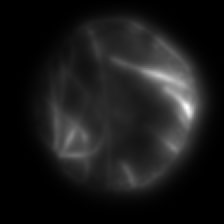}
\end{minipage}

\begin{minipage}{\textwidth}
\includegraphics[width=0.12\textwidth]{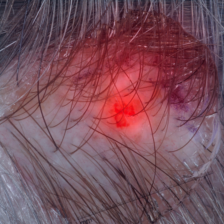}
\includegraphics[width=0.12\textwidth]{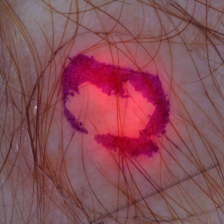}
\includegraphics[width=0.12\textwidth]{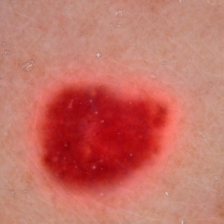}
\includegraphics[width=0.12\textwidth]{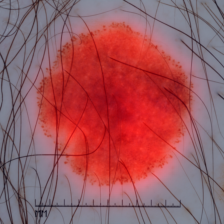}
\includegraphics[width=0.12\textwidth]{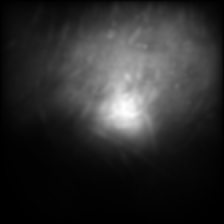}
\includegraphics[width=0.12\textwidth]{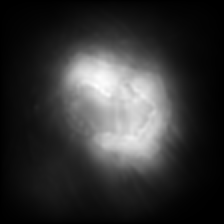}
\includegraphics[width=0.12\textwidth]{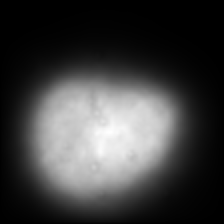}
\includegraphics[width=0.12\textwidth]{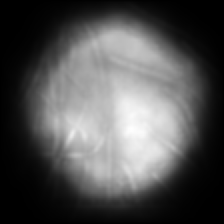}

\includegraphics[width=0.12\textwidth]{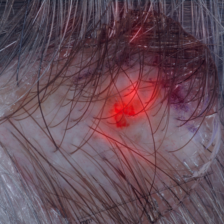}
\includegraphics[width=0.12\textwidth]{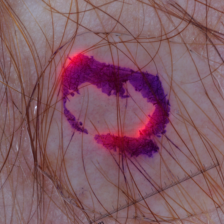}
\includegraphics[width=0.12\textwidth]{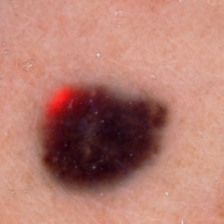}
\includegraphics[width=0.12\textwidth]{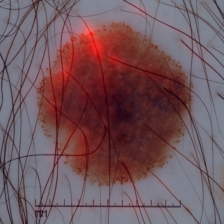}
\includegraphics[width=0.12\textwidth]{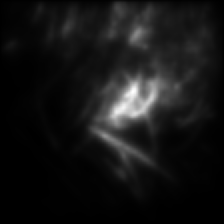}
\includegraphics[width=0.12\textwidth]{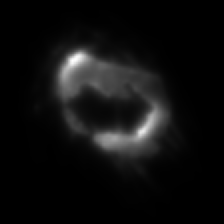}
\includegraphics[width=0.12\textwidth]{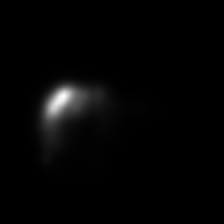}
\includegraphics[width=0.12\textwidth]{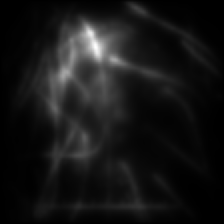}
\end{minipage}
\vspace{-0.1in}
\caption{\label{fig:mel_images} \textbf{First row}: Exemplary input images with their corresponding segmentation masks. \textbf{Following rows}: visualization masks (\textbf{right four images}) and the visualization masks in red overlaid on the input image (\textbf{left four images}) obtained with regular training \textbf{(second row)}, Added seg. mask (\textbf{third row}), Seg. mole (\textbf{fourth row}), Full focus (\textbf{fifth row}), and Half focus (\textbf{sixth row}).}
\vspace{-0.15in}
\end{figure}

In this problem we focus on the classification of dermoscopic images into three unique diagnoses, i.e. Melanoma (type of skin cancer), Nevus (benign lesion), and Seborrheic Keratosis (benign skin tumour). We use images of lesions obtained from a combination of open source data sets, the ISIC Archives \cite{ISIC} ($803$ Melanoma cases, $2107$ Nevus cases, and $288$ Seborrheic Keratosis cases), the Edinburgh Dermofit Image Library \cite{Dermofit} ($76$ Melanoma cases, $331$ Nevus cases, and $257$ Seborrheic Keratosis cases) and the PH$^{2}$ data set \cite{PH2} ($40$ Melanoma cases and $80$ Nevus cases). Additionally in order to balance the data, we generated $350$ images of Melanoma and $750$ images of Seborrheic Keratosis, which were the two classes heavily under-represented in the data set compared to a much larger nevus class. We finally augment the data in the two considered classes by performing horizontal flipping of images such that the class sizes increase to $2685$ for Melanoma and $2772$ for Seborrheic Keratosis. We then augment the entire data set using vertical flipping and random cropping to increase the data set further $6$ times. The final training data set consisted of $16110$ Melanoma cases, $17928$ Nevus cases, and $16632$ Seborrheic Keratosis cases.
Finally, the validation set ($30$ melanoma cases, $78$ nevus cases, and $42$ seborrheic keratosis cases) and test set ($117$ melanoma cases, $393$ nevus cases, and $90$ seborrheic keratosis cases) were obtained from the ISIC data base. We trained a ResNet-50 model using SGD with a learning rate of $1e-4$ and a decay factor of $0.1$ applied every $30$ iterations. We call this a regular training scheme. In addition, we ran two separate experiments to incorporate privileged information: the first one involves concatenating the segmentation mask as the fourth channel (we call this approach \textit{Added seg. mask} in  Table~\ref{tab:melanoma}) and the second one involves masking the input image using segmentation mask to remove background information (we call this approach \textit{Seg. Mole} in  Table~\ref{tab:melanoma}). The background in skin lesion images often contain markers such as hairs and rulers to measure the diameter of the lesion. These sharp features make the training procedure highly prone to over-fitting. The receiver operating characteristics (ROC) obtained on the test data-set are shown in the Table~\ref{tab:melanoma} (the table reports metrics from the ISIC 2017 challenge~\cite{ISIC}) and Figure~\ref{fig:roc_curves} revealing the superiority of both proposed approaches. The exemplary images are captured in Figure~\ref{fig:mel_images} and reveal that the proposed approach focuses network on the desired part of the image preventing making prediction based on the objects in the background, i.e. hairs and rulers. The Full focus method well concentrates the attention of the network that forms prediction on the mole and the surrounding area (consistently with the segmentation mask) and Half focus method obtains the sharpest visualization masks.

\section{Conclusions}
\label{sec:Conclusions}
\vspace{-0.05in}
This paper proposes a methodology for incorporating a privileged information into the training of CNN models by utilizing the visualization technique to guide the network towards specific part of the input image when forming the prediction. The method is easy to implement and computationally cheap as it incurs fewer extra computations than forward pass through the network. Empirically, we find that the method outperforms common baselines on a variety of learning problems.

\clearpage
\newpage
\small{
\bibliographystyle{unsrt}
\bibliography{Double_Super}
}
\end{document}